\pdfoutput=1

\documentclass[11pt]{article}

\usepackage[svgnames]{xcolor}
\usepackage{amsmath,amsthm,amssymb}
\usepackage{mathrsfs}
\usepackage{graphicx}
\usepackage{tabularx}
\usepackage{multirow}
\usepackage{booktabs}
\usepackage{float}
\usepackage{makecell}
\usepackage{colortbl}
\usepackage{algorithm,algorithmic}
\usepackage{subfigure}

\usepackage{ACL2023}

\usepackage{times}
\usepackage{latexsym}

\usepackage[T1]{fontenc}

\usepackage[utf8]{inputenc}

\usepackage{microtype}

\usepackage{inconsolata}

\title{An Empirical Analysis of Parameter-Efficient Methods \\ for Debiasing Pre-Trained Language Models}

\author{First Author \\
  Affiliation / Address line 1 \\
  Affiliation / Address line 2 \\
  Affiliation / Address line 3 \\
  \texttt{email@domain} \\\And
  Second Author \\
  Affiliation / Address line 1 \\
  Affiliation / Address line 2 \\
  Affiliation / Address line 3 \\
  \texttt{email@domain} \\}

\author{Zhongbin Xie$^{1}$, Thomas Lukasiewicz$^{2,1}$ \\
         $^{1}$\,University of Oxford, UK \ $^{2}$\,Vienna University of Technology, Austria \\
         \texttt{zhongbin.xie@cs.ox.ac.uk}, \texttt{thomas.lukasiewicz@tuwien.ac.at}
}

\begin{document}
\maketitle
\begin{abstract}
The increasingly large size of modern pre-trained language models not only makes them inherit more human-like biases from the training corpora, but also makes it computationally expensive to mitigate such biases. In this paper, we investigate recent parameter-efficient methods in combination with counterfactual data augmentation (CDA) for bias mitigation. We conduct extensive experiments with prefix tuning, prompt tuning, and adapter tuning on different language models and bias types to evaluate their debiasing performance and abilities to preserve the internal knowledge of a pre-trained model. 
We find that the parameter-efficient methods (\romannumeral1) are effective in mitigating gender bias, where adapter tuning is consistently the most effective one and prompt tuning is more suitable for GPT-2 than BERT, (\romannumeral2) are less effective when it comes to racial and religious bias, which may be attributed to the limitations of CDA, and  (\romannumeral3) can perform similarly to or sometimes better than full fine-tuning with improved time and memory efficiency, as well as maintain the internal knowledge in BERT and GPT-2, evaluated via fact retrieval and downstream fine-tuning.
\end{abstract}

\section{Introduction}
Pre-trained language models are able to encode rich linguistic and factual knowledge by learning the co-occurrence information of words in large real-world corpora~\citep
{devlin-etal-2019-bert,petroni-etal-2019-language,raffle-etal-2020-jmlr,brown-etal-2020-nips}. Since most of these corpora are internet-based and not carefully curated, they are likely to contain unbalanced or stereotyped information for certain demographic groups. As a result, pre-trained language models are often demonstrated to inherit bias from human society and exhibit potential harms~\citep{blodgett-etal-2020-language,bender-etal-2021-facct,may-etal-2019-measuring,zhao-etal-2019-gender,sheng-etal-2019-woman,nangia-etal-2020-crows,nadeem-etal-2021-stereoset}. Hence, much research effort has been devoted to debias pre-trained language models~\citep{meade-etal-2022-empirical}.

With the size of language models becoming incredibly large~\citep{brown-etal-2020-nips,chinchilla-paper,megatron-turing-nlg-paper}, they are not only at higher risk of exhibiting biased behaviors~\citep{bender-etal-2021-facct}, but also hard to debias because of prohibitive computational cost. Therefore, recent parameter-efficient methods~\citep{he-etal-2022-iclr,ding-etal-2022-arxiv} have been applied to bias mitigation, where only a small portion of the parameters are updated~\citep{lauscher-etal-2021-sustainable-modular,gira-etal-2022-debiasing}. However, these works are limited in terms of evaluation dimensions, making it unclear how different parameter-efficient methods' performance compare to each other, whether one parameter-efficient method is effective across different types of language models, and whether they are also effective for mitigating religious and racial bias in addition to gender bias.
Moreover, direct comparisons with strong post-hoc debiasing methods~\citep{liang-etal-2020-towards,schick-etal-2021-self}, as well as evaluations of bias mitigation's impact on the language model's internal knowledge, are often insufficient.

Given these observations, we investigate three popular parameter-efficient methods, i.e., prefix tuning~\citep{li-liang-2021-prefix}, prompt tuning~\citep{lester-etal-2021-power}, and adapter tuning \citep{houlsby-etal-2019-icml}, in combination with counterfactual data augmentation~\citep[CDA,][]{zhao-etal-2018-gender,zmigrod-etal-2019-counterfactual,webster-etal-2020-arxiv} to debias pre-trained language models. We conduct extensive experiments to study the parameter-efficient methods' performance on two types of language models (BERT~\citep{devlin-etal-2019-bert} for masked language models and GPT-2~\citep{gpt2-paper} for autoregressive language models), three types of social biases (gender, race, and religion), and four types of performance measures (debiasing performance on CrowS-Pairs~\citep{nangia-etal-2020-crows} and StereoSet~\citep{nadeem-etal-2021-stereoset}, language modeling performance on WikiText-2 \citep{merity-etal-2017-iclr} and StereoSet~\citep{nadeem-etal-2021-stereoset}, fact retrieval performance on LAMA~\citep{petroni-etal-2019-language}, as well as downstream fine-tuning performance on WinoBias~\citep{zhao-etal-2018-gender}). We empirically compare to the performance of full fine-tuning and two post-hoc debiasing methods (SentenceDebias~\citep{liang-etal-2020-towards} and SelfDebias~\citep{schick-etal-2021-self}), aiming to comprehensively study the effectiveness of parameter-efficient methods for bias mitigation.\footnote{The code of this paper is available at \url{https://github.com/x-zb/pedb}.}

Our main findings are as follows: 
\begin{itemize}
    \item The parameter-efficient methods are effective in mitigating gender bias. Within the three parameter-efficient methods, adapter tuning is consistently the most effective one for mitigating bias across different types of language models, while prompt tuning is more suitable for GPT-2 than BERT. Comparing to strong post-hoc debiasing methods, parameter-efficient methods are better at preserving the language modeling ability, while still achieving a competitive and sometimes superior debiasing performance.
    \item The parameter-efficient methods are less effective when it comes to mitigating racial and religious bias, where the post-hoc debiasing methods could achieve a more favorable overall performance.
    \item The parameter-efficient methods can perform similarly to or sometimes better than full fine-tuning, with improved time and memory efficiency.   
    \item The parameter-efficient methods can largely maintain the internal knowledge in both BERT and GPT-2, with the reduction in Precision@10 ranging from 0 to 6.8\% across all the LAMA datasets when compared to the original pre-trained model, and with the reduction in average $F_{1}$ scores less than 3.3\% on the hard type-1 examples of WinoBias when compared to full fine-tuning.
\end{itemize}

\section{Parameter-Efficient Methods}
\label{sec:petuning}

In this section, we briefly review three popular parameter-efficient methods investigated in our study: prefix tuning~\citep{li-liang-2021-prefix}, prompt tuning~\citep{lester-etal-2021-power}, and adapter tuning~\citep{pfeiffer-etal-2021-adapterfusion}. In contrast to traditional full fine-tuning where all the model parameters are updated during training, these parameter-efficient methods introduce a small number of \textit{extra} tunable parameters $\varphi$ on top of a frozen pre-trained language model.

Pre-trained language models usually adopt the transformer architecture~\citep{attention-paper} consisting of multiple stacked layers. Assume that there are $N_{layer}$ layers, and $H_{0}^{(i)}\in\mathbb{R}^{T\times d}$ is the input to the $i$-th layer, where $T$ is the sequence length, and $d$ is the model dimension. Then, $H_{0}^{(i)}$ is transformed by the following equations to obtain the output of the $i$-th layer $H_{5}^{(i)}$, which is in turn adopted as the input for the $(i+1)$-th layer:
\begin{align}
    & H_{1,h}^{(i)} = \text{Attn}(H_{0}^{(i)}W_{Q,h}^{(i)},H_{0}^{(i)}W_{K,h}^{(i)},H_{0}^{(i)}W_{V,h}^{(i)}),\notag \\
    & \qquad\qquad\qquad\qquad h=1,2,\dots,N_{head}, \\
    & H_{2}^{(i)} = [H_{1,1}^{(i)};\dots;H_{1,N_{head}}^{(i)}]W_{O}^{(i)}, \\
    & H_{3}^{(i)} = \text{LayerNorm}(H_{0}^{(i)}+H_{2}^{(i)}), \\
    & H_{4}^{(i)} = \text{ReLU}(H_{3}^{(i)}W_{1}^{(i)}+b_{1}^{(i)})W_{2}^{(i)}+b_{2}^{(i)}, \\
    & H_{5}^{(i)} = \text{LayerNorm}(H_{3}^{(i)}+H_{4}^{(i)}).
\end{align}

Here, Eqs.~(1) and (2) constitute the multi-head attention sublayer, where $W_{Q,h}^{(i)}$, $W_{K,h}^{(i)}$, and $W_{V,h}^{(i)}$ denote the projection matrix for the query, key, and value of the $h$-th attention head, respectively; $N_{head}$ is the number of attention heads, and $H_{1,h}^{(i)}\in\mathbb{R}^{T\times (d/N_{head})}$. Eq.~(4) denotes the feed-forward sublayer. $[;]$ denotes the concatenation operation. $H_{j}^{(i)}\in\mathbb{R}^{T\times d}$ for $j=0,2,3,4,5$. The input to the $1$st layer is the embeddings of the input tokens $H_{0}^{(1)}=X\in\mathbb{R}^{T\times d}$.

\paragraph{Prefix tuning.}~\citet{li-liang-2021-prefix} prepend $l$ tunable prefix vectors to the key vectors ($H_{0}^{(i)}W_{K,h}^{(i)}$) and value vectors ($H_{0}^{(i)}W_{V,h}^{(i)}$) of the attention function in Eq.~(1) for each layer:
\begin{align}
    & H_{1,h}^{(i)} = \text{Attn}(H_{0}^{(i)}W_{Q,h}^{(i)},[P_{K,h}^{(i)};H_{0}^{(i)}W_{K,h}^{(i)}], \notag\\
    & \quad [P_{V,h}^{(i)};H_{0}^{(i)}W_{V,h}^{(i)}]), \ h=1,2,\dots,N_{head}.
\end{align}

Here, $P_{K,h}^{(i)},P_{V,h}^{(i)}\!\in\!\mathbb{R}^{l\times(d/N_{head})}$ denote the tunable prefix vectors, and the total tunable parameters are $\varphi\!=\!\{P_{K,h}^{(i)},P_{K,h}^{(i)}\ |\ h\!=\!1,2,\dots,N_{head}, i\!=\!1,2,\dots,N_{layer}\}$.

\paragraph{Prompt tuning.}~\citet{lester-etal-2021-power} prepend $l$ tunable prompt vectors (continuous tokens) only to the input embeddings ($X$), and compute the activations of these prompt vectors in the subsequent layers using the pre-trained transformer's parameters. So, the only modification is:
\begin{equation}
    H_{0}^{(1)}=[P;X]\in\mathbb{R}^{(l+T)\times d},
\end{equation}
where $P\in\mathbb{R}^{l\times d}$ denotes the tunable prompt vectors, and $\varphi=\{P\}$.

\paragraph{Adapter tuning.}~\citet{houlsby-etal-2019-icml} insert the following adapter module between the transformer's sublayers: 
\begin{equation}
     H_{j}^{(i)} \gets  H_{j}^{(i)}+f( H_{j}^{(i)}W_{down}^{(i)})W_{up}^{(i)},
     \label{e:adapter}
\end{equation}
where the intermediate activations $H_{j}^{(i)}$ are first down-projected by $W_{down}^{(i)}\in\mathbb{R}^{d\times(d/r)}$ to a lower dimension $d/r$, and then up-projected back by $W_{up}^{(i)}\in\mathbb{R}^{(d/r)\times d}$ to the model dimension $d$. The adapter also contains a non-linear function $f$ and a residual connection. The hyperparameter $r$ is called the \textit{reduction factor}, which determines the bottleneck dimension $d/r$ and controls the trade-off between parameter efficiency and model capacity.

In our implementation, we adopt \citet{pfeiffer-etal-2021-adapterfusion}'s setting where only a single adapter is inserted after the feed-forward sublayer, since it is found to be the optimal setting among other alternatives~\citep{pfeiffer-etal-2021-adapterfusion}. Thus, all the tunable parameters are $\varphi=\{W_{down}^{(i)},W_{up}^{(i)} |\ i\!=\!1,2,\dots,N_{layer} \}$.\footnote{\citet{pfeiffer-etal-2021-adapterfusion} also insert an additional ``add \& layer norm'' sublayer before the adapter module, so the actual number of tunable parameters is a bit larger.}

\section{Parameter-Efficient Debiasing through Counterfactual Data Augmentation}

We adopt counterfactual data augmentation~\citep[CDA,][]{zhao-etal-2018-gender,zmigrod-etal-2019-counterfactual,webster-etal-2020-arxiv} as our debiasing method to work together with parameter-efficient tuning methods. Since the encoded biases in pre-trained language models originate from the unbalanced training corpora, it is natural to mitigate these biases by re-balancing the training corpora. For example, when we want to mitigate gender bias between the male and female demographic group and encounter the training sentence ``\textit{He is a doctor.}'', CDA would substitute the bias attribute word ``\textit{He}'' with its counterpart ``\textit{She}'' to obtain an additional training sentence ``\textit{She is a doctor.}'', so that both gender groups would have 
equal association with the gender-neutral word ``\textit{doctor}''. Once we have a list of bias attribute words like \{(\textit{he}, \textit{she}), (\textit{man}, \textit{woman}), (\textit{husband}, \textit{wife}), \dots\}, we could retrieve all the occurrences of these bias attribute words in the training corpus, and substitute all of them with their counterparts. 

\begin{algorithm}[t]
    \centering
    \renewcommand{\algorithmicrequire}{\textbf{Input:}}
    \renewcommand{\algorithmicensure}{\textbf{Output:}}
    \caption{Counterfactual Data Augmentation}
    \label{alg:1}
    \begin{algorithmic}[1]
        \REQUIRE original corpus $\mathcal{D}_{0}$, \# demographic groups $N$, \# samples $S (\le N-1)$, bias attribute word list $\{(w_{1}^{(i)},\dots,w_{N}^{(i)})\}_{i=1}^{M}$
        \ENSURE augmented corpus $\mathcal{D}_{1}$
        \STATE $\mathcal{D}_{1} \gets \varnothing$
        \FOR{text sequence $x \in \mathcal{D}_{0}$}
            \STATE Identify the number of demographic groups $n(\le N)$ contained in $x$
            \IF{$n>0$}
                \STATE Generate all the permutations of $N$ demographic groups considered $n$ demographic groups at a time: $\Pi\!=\!\{\pi_{j}\}_{j=1}^{P_{N}^{n}}$, where $\pi_{j}\!=\!(g_{1},\dots,g_{n}),  \{g_{1},\dots,g_{n}\}\!\subset\!\{1,\dots,N\}$
                \IF {$n=N$ and $(1,2,\dots,N) \in \Pi$}
                    \STATE $\Pi \gets \Pi \setminus \{(1,2,\dots,N)\}$
                \ENDIF
                \STATE Sample w/o replacement $S$ permutations $\Pi_{S}=\{\pi_{s}\}_{s=1}^{S}$ from $\Pi$
                \FOR{$\pi_{s} \in \Pi_{S}$}
                    \STATE $x_s\!\gets\!$ Substitute all bias attribute words $w_{k}^{(i)}$ contained in $x$ with $w_{\pi_{s}[k]}^{(i)}$
                    \STATE $\mathcal{D}_{1} \gets \mathcal{D}_{1} \cup \{x_{s}\}$
                \ENDFOR
                \STATE $\mathcal{D}_{1} \gets \mathcal{D}_{1} \cup \{x\}$
            \ENDIF
        \ENDFOR
    \end{algorithmic}
\end{algorithm}

For religious and racial bias where more than two demographic groups are considered, we need to maintain two key properties: (\romannumeral1) we should guarantee \textit{consistency}, 
i.e., we should avoid the case where some occurrences of the bias attribute words in group A are substituted with those in group B, while the other occurrences of (possibly different) bias attribute words in group A are substituted with those in group C, and  (\romannumeral2) we should avoid \textit{collisions}, 
{i.e.}, we should avoid the case where both groups A and B are substituted with group C.
To this end, we should not consider each group independently and adopt random substitution. Rather, we should substitute according to permutations of all the occurred demographic groups in a sentence. Our complete CDA method is formally summarized in Algorithm~\ref{alg:1}.

Note that in Algorithm~\ref{alg:1}, for convenience, we propose to sample a fixed number ($S$) of substitutions for each sentence. This is because the number of possible substitutions ($P_{N}^{n}-1$) for each sentence may vary when the number of occurred demographic groups ($n$) in the sentence varies. In practice, we adopt $N=3$ and $S=2$ for religious and racial bias.

Finally, the parameter-efficient debiasing framework works as follows: we first use Algorithm~\ref{alg:1} to augment an original corpus $\mathcal{D}_{0}$ and obtain the debiasing corpus $\mathcal{D}_{1}$; next,  we use the parameter-efficient tuning methods from Section~\ref{sec:petuning} to solve the following optimization problem:
\begin{equation}
    \min_{\varphi}{\mathcal{L}(\theta_{0},\varphi;\mathcal{D}_{1})},
    \label{equ:opt}
\end{equation}
where $\mathcal{L}$ is either the masked language modeling loss~\citep{devlin-etal-2019-bert} or causal language modeling loss~\citep{gpt2-paper}, $\theta_{0}$ denotes the frozen parameters in the pre-trained language model,
and $\varphi$ denotes the tunable parameters defined in Section~\ref{sec:petuning}.

\section{Conceptual Comparisons with Existing Debiasing Methods}
\label{sec:related}

Most existing debiasing methods are \textit{training-based}, where they introduce a specific debiasing loss to fine-tune a pre-trained model on certain balanced debiasing corpora~\citep{kaneko-bollegala-2021-debiasing,garimella-etal-2021-intelligent,ravfogel-etal-2020-null,cheng-etal-2021-iclr,guo-etal-2022-auto}. These methods are, in general, orthogonal to our parameter-efficient debiasing framework in that we could substitute the (masked) language modeling loss in Eq.~(\ref{equ:opt}) with their specific debiasing loss. In this paper, we only focus on the simple language modeling loss, and leave other kinds of debiasing loss for future work. 

Another important line of debiasing methods applies \textit{post-hoc} mathematical operations on the frozen representations of a language model, such as SentenceDebias~\citep{liang-etal-2020-towards} and SelfDebias~\citep{schick-etal-2021-self}. We briefly review these methods below and make empirical comparisons to parameter-efficient debiasing methods in Section~\ref{sec:exp}. 

\paragraph{SentenceDebias.}~\citet{liang-etal-2020-towards} assume that there is a linear subspace that can capture demographic information in the embedding space, thus trying to identify and remove the demographic information via linear algebra operations.
Specifically, they first leverage a procedure similar to CDA to extract and augment sentences containing bias attribute words from a source corpus. Then, they encode the sentences to embeddings with a pre-trained language model, and obtain a set of difference vectors between the embeddings of sentences in different demographic groups. Next, they perform principle component analysis on the set of difference vectors, and use the first K components to expand a bias subspace. Once the bias subspace is identified, we could debias a new sentence embedding by subtracting its projection on the bias subspace.

\paragraph{SelfDebias.}~\citet{schick-etal-2021-self} assume that a pre-trained language model has a self-diagnosis ability, which can be used to adjust the output probabilities over the vocabulary during language generation.
Specifically, SelfDebias relies on hand-crafted descriptions for each type of bias. It first puts the bias description and the currently generated sentence into a self-diagnosis template, which encourages the language model to generate biased words for the next time step. Then, the probabilities of these detected biased words are scaled down in the actual generation process.

Although no training is needed for these post-hoc debiasing methods, their strong assumptions about bias may harm the language modeling ability of a language model. On the contrary, CDA-based parameter-efficient methods adhere to the original language modeling loss without additional assumptions, which may largely reserve the language modeling ability. Another advantage of CDA-based parameter-efficient methods is that nearly no additional computation is required during inference.

\section{Experiments on Bias Mitigation}
\label{sec:exp}

\subsection{Experimental Setup}

\paragraph{Datasets.} To measure gender, religious, and racial bias in pre-trained language models, we adopt two crowd-sourced datasets: CrowS-Pairs~\citep{nangia-etal-2020-crows} and StereoSet~\citep{nadeem-etal-2021-stereoset}. CrowS-Pairs 
consists of pairs of contrasting sentences, where one is more stereotyping than the other. Its gender, religious, and racial subsets contain 262, 105, and 516 examples, respectively. For StereoSet, we adopt its intra-sentence test, where each example consists of a context sentence and three candidate completions corresponding to stereotypical, anti-stereotypical, and unrelated associations, respectively. We again only adopt the gender, religious, and racial subsets, whose sizes are 1026, 623, and 3996, respectively.   

\paragraph{Evaluation Metrics.} Our evaluation protocol follows \citet{meade-etal-2022-empirical}. We adopt the ``stereotype score'', defined as the percentage of examples for which the language model favors the stereotypical association (or the stereotyping sentence) to the anti-stereotypical association (or the less stereotyping sentence), as the measure of bias. An ideal model that is free of the considered bias should achieve a stereotype score of 50\%. To measure the language modeling ability, we adopt the first 10\% of WikiText-2~\citep{merity-etal-2017-iclr} to compute the perplexity (for autoregressive language models) or pseudo-perplexity~\citep[][for masked language models]{salazar-etal-2020-masked}. We also compute the ``language modeling (LM) score''~\citep{nadeem-etal-2021-stereoset} on all the bias subsets of StereoSet as our second measure of language modeling ability. 

\paragraph{Training Details.} We choose to debias BERT \citep{devlin-etal-2019-bert} and GPT-2~\citep{gpt2-paper}, which represent masked language models and autoregressive language models, respectively. Our implementation is based on the Hugging Face Transformers~\citep{wolf-etal-2020-transformers} and Adapter Hub~\citep{pfeiffer-etal-2020-adapterhub}, and the adopted checkpoints are {\tt bert-base-uncased} (109'514'298 parameters) and {\tt gpt2} (124'439'808 parameters). 
We adopt the English Wikipedia 
as our original debiasing corpus\footnote{We also investigate the effect of different debiasing corpora for GPT-2. See Appendix~\ref{app:gpt2} for details.}, and counterfactually augment it using Algorithm~\ref{alg:1}. The adopted bias attribute words for each type of bias are listed in Appendix~\ref{app:words}. Next, we randomly down-sample 20\% of the augmented Wikipedia as our debiasing corpus. All the CDA-based debiasing methods are trained for two epochs on one TITAN RTX GPU with 24 GB memory. We select optimal training hyperparameters according to the language modeling loss on a validation set (we use 5\% of the augmented debiasing corpus for validation), since the language modeling loss on a balanced dataset is a reasonable proxy for both debiasing performance and language modeling ability. We select hyperparameters using the default seed of 42, and re-train the models for four additional times with different random seeds, to account for CrowS-Pairs and StereoSet's sensitivity to pre-training seeds~\citep{aribandi-etal-2021-reliable}.
More details are in Appendix~\ref{app:details}.

\paragraph{Baselines.} We compare the parameter-efficient methods to full fine-tuning, where \textit{all} the parameters of a language model are tuned, 
For post-hoc debiasing methods, we compare to SentenceDebias~\citep{liang-etal-2020-towards} and Self-Debias~\citep{schick-etal-2021-self}, as described in Section~\ref{sec:related}.

\subsection{Mitigating Gender Bias}

\label{sec:gender}

\begin{table*}[h]
    \centering
    \footnotesize
    \begin{tabular}{p{0.19\textwidth}<{\raggedright}p{0.135\textwidth}<{\centering}p{0.135\textwidth}<{\centering}p{0.135\textwidth}<{\centering}p{0.135\textwidth}<{\centering}}
    \toprule
    {Gender Bias}&{CrowS-Pairs Stereotype Score}&{StereoSet Stereotype Score}&{WikiText2 Perplexity $(\downarrow)$}&{StereoSet LM Score $(\uparrow)$} \\
    \hline
    \hline
    {BERT} & {57.25} & {60.28} & {5.167} &  {84.17} \\
    \hline
    {+Full Fine-Tune} & {56.11{\scriptsize$\pm$2.15}} & {56.43{\scriptsize$\pm$0.72}$^*$} & {5.517{\scriptsize$\pm$0.080}} & 84.22{\scriptsize$\pm$0.19} \\
    {+Prefix Tune ($l\!=\!16$)} & {53.59{\scriptsize$\pm$0.19}$^*$} & {57.82{\scriptsize$\pm$0.46}$^*$} & {\textbf{4.425}{\scriptsize$\pm$0.015}} & {84.75{\scriptsize$\pm$0.15}} \\
    {+Prompt Tune ($l\!=\!16$)} & {57.56{\scriptsize$\pm$1.41}} & {58.07{\scriptsize$\pm$0.60}$^*$} & {4.641{\scriptsize$\pm$0.033}} & {84.71{\scriptsize$\pm$0.16}} \\
    {+Adapter Tune ($r\!=\!48$)} & {\textbf{51.68}{\scriptsize$\pm$0.52}$^{**}$} & {\textbf{56.04}{\scriptsize$\pm$0.43}$^{**}$} & {4.931{\scriptsize$\pm$0.043}} & {\textbf{84.97}{\scriptsize$\pm$0.14}} \\
    \hline
    {+SentenceDebias} & {52.29} & {59.37} & {5.181} & {84.20} \\
    {+SelfDebias} & {52.29} & {59.34} & {7.070} & {84.09} \\
    \hline
    \hline
    {GPT-2} & {56.87} & {62.65} & {29.669} & {91.01} \\
    \hline
    {+Full Fine-Tune} & {55.88{\scriptsize$\pm$1.27}} & {61.88{\scriptsize$\pm$0.55}$^*$} & {81.778{\scriptsize$\pm$0.655}} & {90.24{\scriptsize$\pm$0.14}} \\
    {+Prefix Tune ($l\!=\!16$)} & {54.73{\scriptsize$\pm$0.66}$^*$} & {61.35{\scriptsize$\pm$0.60}$^*$} & {31.400{\scriptsize$\pm$0.108}} & {91.24{\scriptsize$\pm$0.07}} \\
    {+Prompt Tune ($l\!=\!16$)} & {54.12{\scriptsize$\pm$1.14}$^*$} & {61.30{\scriptsize$\pm$0.43}$^*$} & {\textbf{30.630}{\scriptsize$\pm$0.099}} & {\textbf{91.37}{\scriptsize$\pm$0.08}} \\
    {+Adapter Tune ($r\!=\!48$)} & {\textbf{52.29}{\scriptsize$\pm$1.13}$^{**}$} & {60.33{\scriptsize$\pm$0.46}$^{**}$} & {35.255{\scriptsize$\pm$0.345}} & {90.87{\scriptsize$\pm$0.11}} \\
    \hline
    {+SentenceDebias} & {56.11} &  \textbf{56.05} & {56.891} & {87.43} \\
    {+SelfDebias} & {56.11} & {60.84} & {31.482} & {89.07} \\
    \bottomrule
    \end{tabular}
    \caption{Results on mitigating gender bias. For CrowS-Pairs and StereoSet, stereotype scores closer to 50 indicate less bias; for perplexity, lower values are better; for StereoSet LM score, higher values are better. 
    For the CDA-based methods, we report mean$\pm$std from five runs. The best score of all the debiasing methods for each metric is marked in \textbf{bold}.
    $*$:~the reduction in stereotype score w.r.t. that of the original BERT/GPT-2 is statistically significant~($p<0.05$). $**$:~the stereotype score of adapter tuning is significantly ($p<0.05$) lower than those of the other parameter-efficient methods. 
    }
    \label{tab:gender}
\end{table*}

For experiments on mitigating gender bias, we adopt a default reduction factor of $r=48$ in adapter tuning, leading to 304'320 tunable parameters,  which are less than 0.3\% of all the parameters in BERT (109'514'298) or GPT-2 (124'439'808). For prefix tuning, we adopt a prefix length of $l=16$ to obtain a similar amount of tunable parameters (294'912) to adapter tuning. Obtaining a similar amount of tunable parameters for prompt tuning would require an exceptionally large prompt length, even approaching the maximum acceptable sequence length of the pre-trained language models. Therefore, we only set the prompt length $l=16$ (which corresponds to 12'288 tunable parameters) to compare with prefix tuning under the same number of prepending tokens. Evaluation results are shown in Table~\ref{tab:gender}.\footnote{Since SelfDebias preserves 32 tokens for its prefix templates, when measuring perplexity for all the methods in Table~\ref{tab:gender}, the input sequence length is set to 480 (512-32) for BERT and 992 (1024-32) for GPT-2.} 
 
In general, the parameter-efficient methods are effective in reducing stereotype scores, and the reductions are statistically significant ($p<0.05$) under a permutation test~\citep{permutation-test}. 

\smallskip 
\noindent \textbf{Among the three parameter-efficient methods, adapter tuning achieves the best debiasing performance on both CrowS-Pairs and StereoSet, for both BERT and GPT-2.} This demonstrates adapter tuning to be a reliable parameter-efficient method for bias mitigation across different types of language models. Note that our results are also consistent with \citet{he-etal-2022-iclr}'s finding that modifying transformer representations at the feed-forward sublayers (adapter tuning) is more effective than modifying those at the multi-head attention sublayers (prefix tuning).

\smallskip 
\noindent \textbf{Prompt tuning is more effective on GPT-2 than BERT.} Prompt tuning is ineffective in reducing the CrowS-Pairs stereotype score on BERT, but can successfully reduce it on GPT-2, where it even achieves a similar debiasing performance to prefix tuning. This is remarkable given that prompt tuning has much less tunable parameters than prefix tuning. 
This is also consistent with prompt tuning being more effective when T5~\citep{raffle-etal-2020-jmlr} is continuously pre-trained with an autoregressive language modeling loss \citep{lester-etal-2021-power}.

\smallskip 
\noindent \textbf{Comparing to post-hoc debiasing methods, pa\-ra\-me\-ter-efficient methods are better at maintaining the language modeling ability while achieving a similar debiasing performance.} Note that post-hoc debiasing methods sometimes significantly worsen the language modeling ability, {e.g.}, a perplexity of 7.070 for SelfDebias on BERT, a perplexity of 56.891, and a LM score of 87.43 for SentenceDebias on GPT-2. Since a completely \textit{random} language model would achieve the perfect stereotype score (50), but is useless as a language model~\citep{nadeem-etal-2021-stereoset}, the degraded language modeling ability of the post-hoc debiasing methods undermines their true effectiveness for bias mitigation. On the contrary, parameter-efficient methods keep the language modeling loss during CDA training, which helps to preserve or even enhance the language modeling ability. 

\smallskip 
\noindent \textbf{Comparing to full fine-tuning, parameter-ef\-fi\-cient methods can achieve a better or similar performance with improved time and memory efficiency.} Since full fine-tuning updates all the parameters of the language model, it is computationally expensive and prone to be overfitting. When debiasing BERT, full fine-tuning consumes around 19 GB memory, while the parameter-efficient methods consume 12\textasciitilde17 GB memory. Training on the debiasing corpus for full fine-tuning lasts around 6 hours, while that for the parameter-efficient methods lasts 4\textasciitilde5 hours. For GPT-2, full fine-tuning consumes around 18 GB memory with the training time being around 7 hours, while the parameter-efficient methods consume 15\textasciitilde16 GB memory and 5 hours of training time.

\subsection{Mitigating Racial and Religious Bias}

\begin{table*}[h]
    \centering
    \footnotesize
    \begin{tabular}{p{0.19\textwidth}<{\raggedright}p{0.135\textwidth}<{\centering}p{0.135\textwidth}<{\centering}p{0.135\textwidth}<{\centering}p{0.135\textwidth}<{\centering}}
        \toprule
            {Racial Bias}&{CrowS-Pairs Stereotype Score}&{StereoSet Stereotype Score}&{WikiText2 Perplexity $(\downarrow)$}&{StereoSet LM Score $(\uparrow)$} \\
            \hline
            \hline
            {BERT} & {62.33} & {57.03} & {4.899} &  {84.17} \\
            \hline
            {+Full Fine-Tune} & {57.65{\scriptsize$\pm$3.61}$^*$} & {57.67{\scriptsize$\pm$0.70}} & {5.291{\scriptsize$\pm$0.064}} & {83.44{\scriptsize$\pm$0.29}} \\
            {+Prefix Tune ($l\!=\!192$)} & {57.44{\scriptsize$\pm$1.90}$^*$} & {56.95{\scriptsize$\pm$0.39}} & {\textbf{4.448}{\scriptsize$\pm$0.008}} & {\textbf{84.35}{\scriptsize$\pm$0.12}} \\
            {+Prompt Tune ($l\!=\!192$)} & {58.25{\scriptsize$\pm$3.90}$^*$} & {58.17{\scriptsize$\pm$0.55}} & {4.572{\scriptsize$\pm$0.019}} & {83.41{\scriptsize$\pm$0.80}} \\
            {+Adapter Tune ($r\!=\!4$)} & {{57.20}{\scriptsize$\pm$4.16}$^*$} & {59.10{\scriptsize$\pm$0.45}} & {4.903{\scriptsize$\pm$0.071}} & {{84.34}{\scriptsize$\pm$0.20}} \\
            \hline
            {+SentenceDebias} & {62.72} & {57.78} & {4.949} & {83.95} \\
            {+SelfDebias} & \textbf{56.70} & {\textbf{54.30}} & {6.187} & {84.24} \\
            \hline
            \hline
            {GPT-2} & {59.69} & {58.90} & {32.712} & {91.01} \\
            \hline
            {+Full Fine-Tune} & {60.04{\scriptsize$\pm$0.48}} & {{56.68}{\scriptsize$\pm$0.37}$^*$} & {41.781{\scriptsize$\pm$0.240}} & {89.44{\scriptsize$\pm$0.05}} \\
            {+Prefix Tune ($l\!=\!384$)} & {59.61{\scriptsize$\pm$0.51}} & {57.53{\scriptsize$\pm$0.23}$^*$} & {35.346{\scriptsize$\pm$0.073}} & {89.48{\scriptsize$\pm$0.08}} \\
            {+Prompt Tune ($l\!=\!384$)} & {58.76{\scriptsize$\pm$0.92}$^*$} & {57.72{\scriptsize$\pm$0.33}$^*$} & {\textbf{33.983}{\scriptsize$\pm$0.266}} & {89.18{\scriptsize$\pm$0.10}} \\
            {+Adapter Tune ($r\!=\!2$)} & {61.28{\scriptsize$\pm$1.27}} & {57.77{\scriptsize$\pm$0.44}$^*$} & {35.818{\scriptsize$\pm$0.304}} & {89.01{\scriptsize$\pm$0.68}} \\
            \hline
            {+SentenceDebias} & {55.43} & \textbf{56.43} & {37.826} & {\textbf{91.38}} \\
            {+SelfDebias} & {\textbf{53.29}} & {57.33} & {34.851} & {89.53} \\
            \bottomrule
            \\
            \toprule
            {Religious Bias}&{CrowS-Pairs Stereotype Score}&{StereoSet Stereotype Score}&{WikiText2 Perplexity $(\downarrow)$}&{StereoSet LM Score $(\uparrow)$} \\
            \hline
            \hline
            {BERT} & {62.86} & {59.70} & {6.172} &  {84.17} \\
            \hline
            {+Full Fine-Tune} & {65.33{\scriptsize$\pm$2.73}} & {60.76{\scriptsize$\pm$1.38}} & {6.762{\scriptsize$\pm$0.059}} & {83.67{\scriptsize$\pm$0.18}} \\
            {+Prefix Tune ($l\!=\!384$)} & {72.76{\scriptsize$\pm$1.55}} & {60.61{\scriptsize$\pm$0.98}} & {\textbf{5.372}{\scriptsize$\pm$0.010}} & {\textbf{85.42}{\scriptsize$\pm$0.09}} \\
            {+Prompt Tune ($l\!=\!384$)} & {83.05{\scriptsize$\pm$1.85}} & {60.07{\scriptsize$\pm$1.12}} & {5.483{\scriptsize$\pm$0.048}} & {83.80{\scriptsize$\pm$0.58}} \\
            {+Adapter Tune ($r\!=\!2$)} & {68.00{\scriptsize$\pm$4.33}} & {58.93{\scriptsize$\pm$1.19}} & {6.135{\scriptsize$\pm$0.019}} & {84.45{\scriptsize$\pm$0.19}} \\
            \hline
            {+SentenceDebias} & {63.81} & {58.73} & {6.185} & {84.26} \\
            {+SelfDebias} & {\textbf{56.19}} & {\textbf{57.26}} & {7.624} & {84.23} \\
            \hline
            \hline
            {GPT-2} & {62.86} & {63.26} & {32.712} & {91.01} \\
            \hline
            {+Full Fine-Tune} & {\textbf{54.86}{\scriptsize$\pm$1.29}$^*$} & {64.36{\scriptsize$\pm$0.81}} & {45.525{\scriptsize$\pm$0.065}} & {90.20{\scriptsize$\pm$0.06}} \\
            {+Prefix Tune ($l\!=\!384$)} & {60.95{\scriptsize$\pm$0.60}$^*$} & {65.16{\scriptsize$\pm$0.56}} & {{35.226}{\scriptsize$\pm$0.073}} & {\textbf{90.95}{\scriptsize$\pm$0.03}} \\
            {+Prompt Tune ($l\!=\!384$)} & {58.29{\scriptsize$\pm$1.52}$^*$} & {64.89{\scriptsize$\pm$1.52}} & {43.177{\scriptsize$\pm$17.750}} & {90.68{\scriptsize$\pm$0.12}} \\
            {+Adapter Tune ($r\!=\!2$)} & {62.10{\scriptsize$\pm$2.72}} & {62.05{\scriptsize$\pm$0.66}$^*$} & {39.732{\scriptsize$\pm$0.695}} & {90.31{\scriptsize$\pm$0.10}} \\
            \hline
            {+SentenceDebias} & {35.24} & {\textbf{59.62}} & {60.204} & {90.53} \\
            {+SelfDebias} & {58.10} & {60.45} & \textbf{35.174} & {89.36} \\
            \bottomrule
\end{tabular}
    \caption{Results on mitigating racial bias (upper table) and religious bias (lower table). For CrowS-Pairs and StereoSet, stereotype scores closer to 50 indicate less bias; for perplexity\protect\footnotemark, lower values are better; for StereoSet LM score, higher values are better. For the CDA-based methods, we report mean$\pm$std from five runs. The best score of all the debiasing methods for each metric is marked in \textbf{bold}. $*$: the reduction in stereotype score w.r.t. that of the original BERT/GPT-2 is statistically significant ($p<0.05$).
    }
    \label{tab:religion and race}
\end{table*}

When mitigating racial and religious bias, we find that a prefix length of $l=16$ (or,  equivalently, a reduction factor of $r=48$ for adapter tuning) is no longer sufficient for successful debiasing. Therefore, we search $l$ in a broader range of \{48, 96, 192, 384\} (and, correspondingly, $r$ in \{16, 8, 4, 2\}). The results are shown in Table~\ref{tab:religion and race}.

In general, the parameter-efficient methods are less effective when it comes to racial and religious bias. Even the previously strongest method, adapter tuning, is ineffective in many cases such as debiasing BERT on the religion subsets of CrowS-Pairs and StereoSet, and GPT-2 on the race subset of CrowS-Pairs. For GPT-2, prompt tuning is consistently effective on the race subsets of both CrowS-Pairs and StereoSet, but cannot obtain a similar performance on StereoSet's religion subset. In three out of the eight debiasing cases, none of the parameter-efficient methods could reduce the stereotype score in a statistically significant way.

\footnotetext{For the race-debiased models, we set the input sequence length to 320 for BERT and 640 for GPT-2; for the religion-debiased models, we set the input sequence length to 128 for BERT and 640 for GPT-2.}

Moreover, SelfDebias exhibits a superior debiasing performance over the parameter-efficient methods, and its language modeling ability does not severely degenerate as in mitigating gender bias. Indeed, when we calculate the $icat$ score~\citep{nadeem-etal-2021-stereoset}, defined as $lms\ *\ min(ss, 100-ss)/50$ ($lms$ stands for the LM score, and $ss$ stands for the stereotype score on StereoSet), to integrate the debiasing performance and language modeling ability, we can clearly see a better overall performance of SelfDebias over adapter tuning (e.g., on StereoSet's religion subset, the $icat$ score of SelfDebias and adapter tuning is $72.00\ vs.\ 69.37$ for BERT, and $70.68\ vs.\ 68.55$ for GPT-2).

The less successful performance of parameter-efficient methods may be attributed to some limitations of the CDA debiasing method. The bias attribute word lists for race and religion are shorter and contain more noise (i.e., words with multiple or ambiguous meanings) than that for gender, which may undermine the diversity and quality of the augmented training corpus. On the contrary, SelfDebias relies on bias descriptions that  contain less noise and could generalize with the help of the language model's own knowledge. Given this analysis, future work could explore how to adopt parameter-efficient methods to debiasing techniques other than CDA to overcome these limitations.

\section{Impact on Internal Knowledge}
\label{sec:impact}

\subsection{Fact Retrieval}

\begin{table*}[h]
    \centering
    \footnotesize
    \begin{tabular}{l|p{0.035\textwidth}<{\centering}p{0.035\textwidth}<{\centering}p{0.035\textwidth}<{\centering}|p{0.035\textwidth}<{\centering}p{0.035\textwidth}<{\centering}p{0.035\textwidth}<{\centering}|p{0.035\textwidth}<{\centering}p{0.035\textwidth}<{\centering}p{0.035\textwidth}<{\centering}|p{0.035\textwidth}<{\centering}p{0.035\textwidth}<{\centering}p{0.035\textwidth}<{\centering}}
    \toprule
    \multirow{2}*{}&\multicolumn{3}{c|}{Google-RE}&\multicolumn{3}{c|}{T-REx}&\multicolumn{3}{c|}{ConceptNet}&\multicolumn{3}{c}{SQuAD} \\
    {} & {P@1} & {P@10} & {MRR} & {P@1} & {P@10} & {MRR} & {P@1} & {P@10} & {MRR} & {P@1} & {P@10} & {MRR} \\
    \midrule
    {BERT} & {9.25} & {28.69} & {15.96} & {29.48} & {56.87} & {38.63} & {15.11} & {38.77} & {23.10} & {13.11} & {44.59} & {23.30} \\
    \hline
    {+Full Fine-Tune} & {7.47} & {21.43} & {12.43} & {26.93} & {\textbf{53.72}} & {35.85} & {14.89} & {\textbf{37.59}} & {22.57} & \cellcolor{LightGreen}{\textbf{14.43}} & \cellcolor{LightGreen}{\textbf{47.21}} & \cellcolor{LightGreen}{\textbf{24.78}}   \\
    {+Prefix Tune ($l\!=\!16$)} & {8.23} & {22.54} & {13.43} & {27.68} & {53.64} & {36.38} & \textbf{15.05} & {37.42} & \textbf{22.73} & {12.79} & \cellcolor{LightGreen}{46.56} & \cellcolor{LightGreen}{23.91}  \\
    {+Prompt Tune ($l\!=\!16$)} & \textbf{8.68} & \textbf{23.19} & {\textbf{14.04}} & \textbf{28.28} & {53.59} & \textbf{36.88} & {14.58} & {36.58} & {22.11} & {12.79} & \cellcolor{LightGreen}{46.89} & \cellcolor{LightGreen}{23.54}  \\
    {+Adapter Tune ($r\!=\!48$)} & {8.51} & {21.97} & {13.39} & {26.92} & {51.65} & {35.27} & {14.75} & {36.47} & {22.13} & {11.80} & {44.26} & {22.59}  \\
    \midrule
    {GPT-2} & {1.51} & {10.88} & {5.04} & {9.36} & {31.10} & {16.78} & {5.91} & {19.01} & {10.42} & {3.15} & {17.48} & {7.53} \\
    \hline
    {+Full Fine-Tune} & \cellcolor{LightGreen}{\textbf{3.40}} & \cellcolor{LightGreen}{\textbf{15.10}} & \cellcolor{LightGreen}{\textbf{7.44}} & {7.76} & \cellcolor{LightGreen}{33.04} & {15.90} & {4.87} & {16.47} & {8.86} & {1.75} & \cellcolor{LightGreen}{\textbf{18.18}} & {6.78}   \\
    {+Prefix Tune ($l\!=\!16$)} & \cellcolor{LightGreen}{2.33} & \cellcolor{LightGreen}{12.56} & \cellcolor{LightGreen}{6.14} & \cellcolor{LightGreen}{\textbf{10.13}} & \cellcolor{LightGreen}{\textbf{33.38}} & \cellcolor{LightGreen}{\textbf{17.98}} & \cellcolor{LightGreen}{\textbf{5.99}} & \cellcolor{LightGreen}{\textbf{19.42}} & \cellcolor{LightGreen}{\textbf{10.53}} & {2.10} & \cellcolor{LightGreen}{17.83} & \cellcolor{LightGreen}{\textbf{7.53}}  \\
    {+Prompt Tune ($l\!=\!16$)} & {1.14} & {9.79} & {4.39} & {8.00} & {30.29} & {15.70} & \cellcolor{LightGreen}{5.95} & \cellcolor{LightGreen}{19.03} & \cellcolor{LightGreen}{\textbf{10.53}} & \textbf{2.45} & {16.78} & {7.16}  \\
    {+Adapter Tune ($r\!=\!48$)} & \cellcolor{LightGreen}{2.49} & \cellcolor{LightGreen}{14.11} & \cellcolor{LightGreen}{6.59} & {9.35} & \cellcolor{LightGreen}{32.61} & \cellcolor{LightGreen}{17.20} & {5.79} & \cellcolor{LightGreen}{19.09} & {10.26} & {2.10} & \cellcolor{LightGreen}{\textbf{18.18}} & {7.03}  \\
    \bottomrule
    \end{tabular}
    \caption{Fact retrieval results of the original and debiased models on the four LAMA datasets. For all the metrics (precision-at-1 (P@1), precision-at-10 (P@10), and mean reciprocal rank (MRR)), higher values are better. For the debiased models, the best score under each metric is in \textbf{bold}, while the scores not worse than those from the original BERT/GPT-2 are highlighted in \colorbox{LightGreen}{green}.}
    \label{tab:lama}
\end{table*}

To investigate the impact of bias mitigation on the factual knowledge encoded in pre-trained language models, we take the gender-debiased models from Section~\ref{sec:gender} and evaluate them on the four LAMA datasets~\citep{petroni-etal-2019-language}.\footnote{Instead of using the intersectional vocabulary of several pre-trained models, as in \citet{petroni-etal-2019-language}, we adopt each pre-trained model's full vocabulary, since we do not aim to compare the performance across different pre-trained models.} The results are shown in Table~\ref{tab:lama}. We report the results from a single run (with the default seed 42) to save computation in Table~\ref{tab:lama} and \ref{tab:winobias}.

\smallskip 
\noindent \textbf{The parameter-efficient methods can largely maintain the factual knowledge of a language model, with the reduction in Precision@10 ranging from 0 to 6.8\% across all the datasets and pre-trained models.} Surprisingly, for BERT on SQuAD and GPT-2 on all the four datasets, quite a number of the results are actually improved. We attribute these improvements to the fact that Wikipedia contains a lot of factual knowledge, and continuously training on it can enhance the internal knowledge of a language model. 

Comparing the performance between full fine-tuning and parameter-efficient tuning, we find that the former performs best on SQuAD with BERT and Google-RE with GPT-2, while the latter performs better in the rest of the settings. In general, the performance gaps are marginal.

\begin{table*}[h]
    \centering
    \footnotesize
    \begin{tabular}{l|cccc|cccc}
    \toprule
    \multirow{2}*{}&\multicolumn{4}{c|}{Type-1}&\multicolumn{4}{c}{Type-2} \\
    {} & {$F_{1\!-\!pro}$} & {$F_{1\!-\!anti}$} & {Avg} & {Diff} & {$F_{1\!-\!pro}$} & {$F_{1\!-\!anti}$} & {Avg} & {Diff} \\
    \hline
    \hline
    {BERT} & {} & {} & {} & {} & {} & {} & {} & {} \\
    {+Full Fine-Tune} & {70.95} & {68.04} & {69.50} & {2.91} & {99.49} & {99.49} & {99.49} & {0} \\
    {+Prefix Tune ($l\!=\!16$)} & {65.08} & {64.57} & {64.83} & {0.51} & {99.49} & {99.49} & {99.49} & {0} \\
    {+Prompt Tune ($l\!=\!16$)} & {56.56} & {53.33} & {54.95} & {3.23} & {99.24} & {99.24} & {99.24} & {0} \\
    {+Adapter Tune ($r\!=\!48$)} & {66.50} & {65.99} & {66.25} & {0.51} & {99.49} & {99.49} & {99.49} & {0} \\
    \hline
    \hline
    {GPT-2} & {} & {} & {} & {} & {} & {} & {} & {} \\
    {+Full Fine-Tune} & {63.33} & {63.47} & {63.40} & {-0.14} & {99.49} & {99.49} & {99.49} & {0} \\
    {+Prefix Tune ($l\!=\!16$)} & {51.66} & {52.79} & {52.23} & {-1.13} & {99.49} & {99.49} & {99.49} & {0} \\
    {+Prompt Tune ($l\!=\!16$)} & {53.46} & {52.36} & {52.91} & {1.10} & {99.24} & {99.24} & {99.24} & {0} \\
    {+Adapter Tune ($r\!=\!48$)} & {60.70} & {59.96} & {60.33} & {0.74} & {99.49} & {99.49} & {99.49} & {0} \\
    \bottomrule
    \end{tabular}
    \caption{Evaluation results on the WinoBias' type-1 and type-2 test sets. We report the $F_1$ score on the pro-stereotypical examples ($F_{1\!-\!pro}$), anti-stereotypical examples ($F_{1\!-\!anti}$), their average (Avg), and their difference (Diff) to measure the models' performance on both the coreference resolution task and the bias mitigation task.}
    \label{tab:winobias}
\end{table*}

\subsection{Downstream Fine-Tuning}

We further investigate the impact of bias mitigation on knowledge transfer to downstream tasks via fine-tuning. Since neural network models suffer from catastrophic forgetting~\citep{french-1999}, a debiased model may forget the encoded knowledge in the original language model, and conversely a fine-tuned model may forget the debiasing knowledge in the debiased model. Therefore, it is important to adopt an evaluation dataset that can simultaneously evaluate downstream task performance and debiasing performance. We choose the coreference resolution dataset WinoBias~\citep{zhao-etal-2018-gender} to fulfill the above requirements. 

We append each example from WinoBias ({e.g.}, \textit{The physician hired the secretary because he was overwhelmed with clients.}) with the suffix ``\{Pronoun\} \textit{refers to the} \{Candidate\}\textit{.}'' (\{Pronoun\} is ``\textit{He}'' in this example), and then measure the probability of the model completing the sentence with different candidates (``\textit{physician}'' and ``\textit{secretary}'' in this example) to determine the coreference result. We adopt both the type-1 and type-2 test sets of WinoBias, where type-1 examples are harder to resolve as they contain no syntactic cues. We adopt WinoBias' dev set to fine-tune an original pre-trained language model using either full fine-tuning or parameter-efficient tuning.\footnote{See Appendix~\ref{app:details} for more details.} The results are shown in Table~\ref{tab:winobias}. 

\smallskip 
\noindent \textbf{On type-1 examples, adapter tuning achieves a comparable performance to full fine-tuning for both BERT and GPT-2, with the reduction in average $F_{1}$ scores less than 3.3\%.} On BERT, adapter tuning achieves a much better debiasing performance (Diff$=0.51$) than full fine-tuning, while on GPT-2 it is slightly more biased. Nevertheless, both of them can be considered effective simultaneously on the coreference resolution task and debiasing task.
The performance gap between full fine-tuning and prefix/prompt tuning is more significant, but the latter can still achieve a nearly perfect performance on the easier type-2 examples. 


\section{Conclusion}
In this study, we investigated the performance of prefix tuning, prompt tuning, and adapter tuning on mitigating social bias and preserving the linguistic and factual knowledge for two types of pre-trained language models. Our results demonstrated the effectiveness and efficacy of parameter-efficient methods in combination with CDA, and also revealed their performance limitations by comparing to post-hoc debiasing methods. 
We hope that our study can make it more accessible for others to debias pre-trained language models with reduced computational requirements, and contribute to fair and inclusive NLP.

\section{Limitations}

Due to the restrictions of the adopted benchmarks and resources, our evaluation bears the following limitations: (\romannumeral1) We only focus on social biases in the English language and North American cultures. This is due to the fact that both CrowS-Pairs and StereoSet are generated by crowd workers from North America. Future work can extend our analysis to other languages and cultures with the corresponding resources such as the French CrowS-Pairs~\citep{neveol-etal-2022-french} and multilingual WEAT~\citep{lauscher-glavas-2019-consistently}. (\romannumeral2) Our evaluation has a limited coverage over different kinds of harms according to \citet{blodgett-etal-2020-language}. CrowS-Pairs, StereoSet, and WinoBias all focus on stereotyping, a kind of representational harm, while others like allocational harms are untouched. Developing methods to measure these harms generally requires in-depth interactions between technologists and customers. 
\citet{blodgett-etal-2021-stereotyping} also point out several conceputalization and operationalization pitfalls in the above three bias benchmarks, which limits the validity of the results evaluated on them.
(\romannumeral3)~Due to the incomplete bias attribute word lists, our CDA-based debiasing method is by no means fair enough to cover all the minority groups ({e.g.}, groups with non-binary genders). Therefore the current debiasing method in this paper can only be used to mitigate bias among the demographic groups mentioned in Appendix~\ref{app:words}. We recommend more complete resources such as the gender-inclusive word list in \citep{cao-daume-iii-2021-toward} for real-world scenarios.


\section*{Acknowledgements}
This work was supported by the Alan Turing Institute under the EPSRC grant EP/N510129/1, by the AXA Research Fund, and by the EU TAILOR grant 952215.
We also acknowledge the use of Oxford’s ARC facility, of the EPSRC-funded Tier 2 facilities JADE (EP/P020275/1), and of GPU computing support by Scan Computers International Ltd. 

\bibliography{anthology,custom}
\bibliographystyle{acl_natbib}

\clearpage
\appendix

\section{Bias Attribute Words}
\label{app:words}

We adopt the same bias attribute words as \citet{meade-etal-2022-empirical}, where the list for gender is from \citet{zhao-etal-2018-gender} and that for religion is from \citet{liang-etal-2020-towards}.

\paragraph{Gender:} (\textit{actor}, \textit{actress}), (\textit{actors}, \textit{actresses}), (\textit{airman}, \textit{airwoman}), (\textit{uncle}, \textit{aunt}), (\textit{uncles}, \textit{aunts}), (\textit{boy}, \textit{girl}), (\textit{boys}, \textit{girls}), (\textit{groom}, \textit{bride}), (\textit{grooms}, \textit{brides}), (\textit{brother}, \textit{sister}), (\textit{brothers}, \textit{sisters}), (\textit{businessman}, \textit{businesswoman}), (\textit{businessmen}, \textit{businesswomen}), (\textit{chairman}, \textit{chairwoman}), (\textit{chairmen}, \textit{chairwomen}), (\textit{dude}, \textit{chick}), (\textit{dudes}, \textit{chicks}), (\textit{dad}, \textit{mom}), (\textit{dads}, \textit{moms}), (\textit{daddy}, \textit{mommy}), (\textit{daddies}, \textit{mommies}), (\textit{son}, \textit{daughter}), (\textit{sons}, \textit{daughters}), (\textit{father}, \textit{mother}), (\textit{fathers}, \textit{mothers}), (\textit{male}, \textit{female}), (\textit{males}, \textit{females}), (\textit{guy}, \textit{gal}), (\textit{guys}, \textit{gals}), (\textit{gentleman}, \textit{lady}), (\textit{gentlemen}, \textit{ladies}), (\textit{grandson}, \textit{granddaughter}), (\textit{grandsons}, \textit{granddaughters}), (\textit{guy}, \textit{girl}), (\textit{guys}, \textit{girls}), (\textit{he}, \textit{she}), (\textit{himself}, \textit{herself}), (\textit{him}, \textit{her}), (\textit{his}, \textit{her}), (\textit{husband}, \textit{wife}), (\textit{husbands}, \textit{wives}), (\textit{king}, \textit{queen}), (\textit{kings}, \textit{queens}), (\textit{lord}, \textit{lady}), (\textit{lords}, \textit{ladies}), (\textit{sir}, \textit{maam}), (\textit{man}, \textit{woman}), (\textit{men}, \textit{women}), (\textit{sir}, \textit{miss}), (\textit{mr.}, \textit{mrs.}), (\textit{mr.}, \textit{ms.}), (\textit{policeman}, \textit{policewoman}), (\textit{prince}, \textit{princess}), (\textit{princes}, \textit{princesses}), (\textit{spokesman}, \textit{spokeswoman}), (\textit{spokesmen}, \textit{spokeswomen})

\paragraph{Religion:} (\textit{jewish}, \textit{christian}, \textit{muslim}), (\textit{jews}, \textit{christians}, \textit{muslims}), (\textit{torah}, \textit{bible}, \textit{quran}), (\textit{synagogue}, \textit{church}, \textit{mosque}), (\textit{rabbi}, \textit{priest}, \textit{imam}), (\textit{judaism}, \textit{christianity}, \textit{islam})

\paragraph{Race:} (\textit{black}, \textit{caucasian}, \textit{asian}), (\textit{african}, \textit{caucasian}, \textit{asian}), (\textit{black}, \textit{white}, \textit{asian}), (\textit{africa}, \textit{america}, \textit{asia}), (\textit{africa}, \textit{america}, \textit{china}), (\textit{africa}, \textit{europe}, \textit{asia})

\section{Additional Training Details}
\label{app:details}

For all the experiments on parameter-efficient tuning methods and full fine-tuning, we use the default settings of the AdamW optimizer~\citep{adamw-paper} and a linear learning rate scheduler from the Hugging Face library. 

For the debiasing experiments trained on Wikipedia, 
we fix the number of training epochs to 2 and greedily search initial learning rate from \{5e-1, 5e-2, 5e-3, 5e-4, 5e-5, 5e-6, 5e-7\} according to the language modeling loss on the validation set (we use 5\% of the augmented debiasing corpus for validation). For experiments trained on WinoBias, we greedily search training epochs from \{10, 20, 30, 50, 100, 200\} and initial learning rate from \{5e-1, 5e-2, 5e-3, 5e-4, 5e-5, 5e-6, 5e-7\} according to the Avg $F_{1}$ score on type-1 examples in the validation set (we use 5\% of the training set for validation). The hyperparameter values to reproduce our results in Sections~\ref{sec:exp} and \ref{sec:impact} are  in Table~\ref{tab:hyperparam}. 

Implementations of SentenceDebias and SelfDebias are based on \citet{meade-etal-2022-empirical}'s, where we also follow their default parameter settings.

\begin{table}[h]
    \centering
    \footnotesize
    \begin{tabular}{p{0.4\columnwidth}<{\raggedright}|p{0.1\columnwidth}<{\centering}p{0.1\columnwidth}<{\centering}p{0.08\columnwidth}<{\centering}}
    \toprule
    {} & {lr} & {epoch} & {bsz} \\
    \hline
    \hline
    \multicolumn{4}{l}{For results in Table~\ref{tab:gender} (gender bias)} \\
    \hline
    {BERT}  \\
    {+Full Fine-Tune} & {5e-5} & {2} & {16} \\
    {+Prefix Tune ($l\!=\!16$)} & {5e-3} & {2} & {16} \\
    {+Prompt Tune ($l\!=\!16$)} & {5e-1} & {2} & {16} \\
    {+Adapter Tune ($r\!=\!48$)} & {5e-4} & {2} & {16} \\
    {GPT-2}  \\
    {+Full Fine-Tune} & {5e-5} & {2} & {8}  \\
    {+Prefix Tune ($l\!=\!16$)} & {5e-3} & {2} & {8} \\
    {+Prompt Tune ($l\!=\!16$)} & {5e-2} & {2} & {8} \\
    {+Adapter Tune ($r\!=\!48$)} & {5e-4} & {2} & {8}  \\
    \hline
    \hline
    \multicolumn{4}{l}{For results in Table~\ref{tab:religion and race}'s upper sub-table (racial bias)} \\
    \hline
    {BERT}  \\
    {+Full Fine-Tune} & {5e-5} & {2} & {16} \\
    {+Prefix Tune ($l\!=\!192$)} & {5e-3} & {2} & {16} \\
    {+Prompt Tune ($l\!=\!192$)} & {5e-3} & {2} & {16} \\
    {+Adapter Tune ($r\!=\!4$)} & {5e-4} & {2} & {16} \\
    {GPT-2}  \\
    {+Full Fine-Tune} & {5e-6} & {2} & {8}  \\
    {+Prefix Tune ($l\!=\!384$)} & {5e-3} & {2} & {8} \\
    {+Prompt Tune ($l\!=\!384$)} & {5e-1} & {2} & {8} \\
    {+Adapter Tune ($r\!=\!2$)} & {5e-3} & {2} & {8}  \\
    \hline
    \hline
    \multicolumn{4}{l}{For results in Table~\ref{tab:religion and race}'s lower sub-table (religious bias)} \\
    \hline
    {BERT}  \\
    {+Full Fine-Tune} & {5e-5} & {2} & {16} \\
    {+Prefix Tune ($l\!=\!384$)} & {5e-3} & {2} & {16} \\
    {+Prompt Tune ($l\!=\!384$)} & {5e-3} & {2} & {16} \\
    {+Adapter Tune ($r\!=\!2$)} & {5e-4} & {2} & {16} \\
    {GPT-2}  \\
    {+Full Fine-Tune} & {5e-6} & {2} & {8}  \\
    {+Prefix Tune ($l\!=\!384$)} & {5e-3} & {2} & {8} \\
    {+Prompt Tune ($l\!=\!384$)} & {5e-1} & {2} & {8} \\
    {+Adapter Tune ($r\!=\!2$)} & {5e-5} & {2} & {8}  \\
    \hline
    \hline
    \multicolumn{4}{l}{For results in Table~\ref{tab:winobias} (WinoBias)} \\
    \hline
    {BERT}  \\
    {+Full Fine-Tune} & {5e-6} & {30} & {16} \\
    {+Prefix Tune ($l\!=\!16$)} & {5e-2} & {20} & {16} \\
    {+Prompt Tune ($l\!=\!16$)} & {5e-1} & {20} & {16} \\
    {+Adapter Tune ($r\!=\!48$)} & {5e-4} & {20} & {16} \\
    {GPT-2}  \\
    {+Full Fine-Tune} & {5e-5} & {20} & {16}  \\
    {+Prefix Tune ($l\!=\!16$)} & {5e-3} & {200} & {16} \\
    {+Prompt Tune ($l\!=\!16$)} & {5e-4} & {100} & {16} \\
    {+Adapter Tune ($r\!=\!48$)} & {5e-4} & {50} & {16}  \\
    \bottomrule
    \end{tabular}
    \caption{Hyperparameter values adopted during training. ``lr'' denotes initial learning rate; ``epoch'' denotes total training epochs; ``bsz'' denotes batch size.}
    \label{tab:hyperparam}
\end{table}

\section{Effect of the Debiasing Corpus for GPT-2}
\label{app:gpt2}

\begin{table*}[htbp]
    \centering
    \footnotesize
    \begin{tabular}{p{0.1\textwidth}<{\raggedright}|p{0.19\textwidth}<{\raggedright}p{0.135\textwidth}<{\centering}p{0.135\textwidth}<{\centering}p{0.135\textwidth}<{\centering}p{0.135\textwidth}<{\centering}}
    \toprule
    {Debiasing Corpus}& {} &{CrowS-Pairs Stereotype Score}&{StereoSet Stereotype Score}&{WikiText2 Perplexity $(\downarrow)$}&{Stereoset LM Score $(\uparrow)$} \\
    \hline
    \hline
    {} & {GPT-2} & {56.87} & {62.65} & {29.669} & {91.01} \\
    \hline
    {Wikipedia} & {+Full Fine-Tune} & {56.87} & {61.30} & {80.499} & {90.23} \\
    {(single} & {+Prefix Tune ($l\!=\!16$)} & {55.34} & {62.02} & {31.567} & {91.14} \\
    {sentence)} & {+Prompt Tune ($l\!=\!16$)} & {52.29} & {60.95} & {30.534} & {91.29} \\
    {} & {+Adapter Tune ($r\!=\!48$)} & {\textbf{51.15}} & \textbf{60.50} & {34.910} & {90.80} \\
    \hline
    \hline
    {Wikipedia} & {+Full Fine-Tune} & {56.49} & {61.74} & {56.527} & {90.19} \\
    {(example} & {+Prefix Tune ($l\!=\!16$)} & {58.40} & {62.67} & {31.935} & {91.22} \\
    {length=1024} & {+Prompt Tune ($l\!=\!16$)} & {56.87} & {63.37} & {32.461} & {91.03} \\
    {tokens)} & {+Adapter Tune ($r\!=\!48$)} & {59.92} & {62.31} & {34.527} & {90.75} \\
    \hline
    \hline
    {OpenWebText} & {+Full Fine-Tune} & {55.73} & {62.43} & {38.252} & {90.60} \\
    {(example} & {+Prefix Tune ($l\!=\!16$)} & {53.44} & {60.94} & {31.592} & {90.31} \\
    {length=1024} & {+Prompt Tune ($l\!=\!16$)} & {53.05} & {62.68} & \textbf{30.464} & \textbf{91.41} \\
    {tokens)} & {+Adapter Tune ($r\!=\!48$)} & {56.87} & {61.94} & {33.130} & {90.87} \\
    \bottomrule
    \end{tabular}
    \caption{Results on gender debiasing and language modeling for GPT-2 using different debiasing corpora.} 
    \label{tab:effect}
\end{table*}

For consistency, we adopt the same debiasing corpus (the English Wikipedia) for both BERT and GPT-2 in Section~\ref{sec:gender}, where each training example consists of a single sentence (the average sentence length in our corpus is around 107 tokens). However, this setting is different from the original pre-training settings of GPT-2~\citep{gpt2-paper} in terms of example length and data source. Therefore, we further investigate debiasing \mbox{GPT-2} on two other debiasing corpora: for one corpus, we still use Wikipedia but concatenate all the sentences into a long sequence and truncate it into examples of 1024 tokens; for the other corpus, we use 1\% of  OpenWebText\footnote{\url{https://skylion007.github.io/OpenWebTextCorpus/}}, which is a public replicate of GPT-2's private pre-training corpus, and truncate it into examples of 1024 tokens. The results are shown in Table~\ref{tab:effect}.\footnote{We report the results from a single run (with the default seed 42) to save computation.} 

Comparing the results on Wikipedia, with single sentence and example length 1024 tokens, in Table~\ref{tab:effect}, we can see that the former is consistently better. This indicates that these methods favor shorter example lengths. We conjecture that this is due to \mbox{GPT-2}'s language modeling objective being an average over all the tokens in an example. Therefore, the counterfactual token's signal will be less significant if it is close to the end of a long example.

Comparing the last two blocks, we can see that the results from the debiasing methods trained on OpenWebText are superior to those trained on Wikipedia under the same example length of 1024. This indicates that using a similar data source to the original pre-training corpus is beneficial. For full fine-tuning, this can improve perplexity to 38.252. For the parameter-efficient methods, the improvements are more significant on stereotype scores. Given that parameter-efficient methods' model capacity is limited, if we allocate some capacity for adapting to new data sources, it is reasonable for the debiasing performance to be negatively affected.

\end{document}